\documentclass[10pt,twocolumn,letterpaper]{article}

\usepackage[pagenumbers]{cvpr} 

\definecolor{cvprblue}{rgb}{0.21,0.49,0.74}
\usepackage[pagebackref,breaklinks,colorlinks,allcolors=cvprblue]{hyperref}

\def\paperID{*****} 
\def\confName{CVPR}
\def\confYear{2026}
\definecolor{safegreen}{HTML}{27AE60}
\definecolor{safered}{HTML}{C0392B}
\definecolor{safeorange}{HTML}{E67E22}
\definecolor{safeblue}{HTML}{2E75B6}
\title{Is VLA Reasoning Faithful? Probing Safety of Chain-of-Causation \\
in Autonomous Driving Models\thanks{Accepted DriveX, CVPR 2026}}

\author{Nicanor Mayumu%
	\thanks{Corresponding author.}%
	\thanks{Email: \texttt{nicanor.mayumu@csu.edu.cn}}%
	\\
	School of Computer Science and Engineering\\
	Central South University\\
	932 Lushan Road, Changsha, Hunan\\
	P. R. China
	\and
	Xiaoheng Deng%
	\thanks{Email: \texttt{dxh@csu.edu.cn}}%
	\\
	School of Computer Science and Engineering\\
	Central South University\\
	932 Lushan Road, Changsha, Hunan\\
	P. R. China
	\and
	Patrick Mukala%
	\thanks{Email: \texttt{pmukala@uow.edu.au}}%
	\\
	School of Computer Science\\
	University of Wollongong in Dubai\\
	UOWD Building, Dubai Knowledge Park\\
	Dubai, United Arab Emirates
}

\begin{document}
\maketitle
\begin{abstract}
	We present the first systematic study of faithfulness in Vision-Language-Action (VLA) driving models, analyzing 300 Alpamayo-R1-10B inferences across 100 diverse PhysicalAI-AV scenarios. Our main finding is that output natural-language rationales with trajectories may be significantly unfaithful: (i) overall reasoning fidelity is only 42.5\%, with Chain-of-Causation matching scene reality less than half the time; (ii) 94 missed pedestrians in one-third of pedestrian-relevant scenes; (iii) 97.7\% trajectory fragility under mild visual perturbations; and (iv) only 48.3\% mean reasoning-action consistency, with 53.3\% of inferences exhibiting low consistency, including 37.9\% of stop-claimed cases where the model continues instead. We formalize faithfulness information-theoretically, define entity and action fidelity with verification criteria, and outline a four-component safety architecture aligned with these results.
\end{abstract}
\section{Introduction}
\label{sec:intro}

Vision-Language-Action (VLA) models for autonomous driving move beyond modular stacks toward end-to-end systems that jointly reason and act~\cite{alpamayo2025}. Models such as Alpamayo~\cite{alpamayo2025}, EMMA~\cite{hwang2024emma}, and DriveVLM~\cite{tian2024drivevlm} emit natural-language Chain-of-Causation (CoC) traces alongside control, which is attractive to regulators and the public who expect interpretability.

Yet work on large language models shows that explanations are often \emph{unfaithful}: they can rationalize decisions post hoc rather than reflect the underlying computation~\cite{turpin2024language, lanham2023measuring}. For driving, this matters urgently: if CoC does not reflect scene understanding or planned behavior, a trace like ``stop for the pedestrian ahead'' can mislead when the pedestrian is missed, or the stop is not executed, arguably worse than silence, because it invites misplaced trust.

VLAs exacerbate this concern structurally: language and trajectory arise from coupled but distinct pathways (e.g., a VLM conditioned on vision versus a trajectory decoder), so there is no built-in guarantee that reasoning and action are causally aligned.

We provide the first systematic study of reasoning faithfulness in a production-scale VLA driving model, with three contributions: (1) a formal fidelity framework based on information theory and counterfactuals, decomposing faithfulness into entity fidelity, action fidelity, and perturbation sensitivity with explicit verification criteria; (2) quantitative evidence of unfaithfulness, 42.5\% overall fidelity, 8.9\% hallucination rate, 94 missed pedestrians, and 53.3\% reasoning-action inconsistency over 300 inferences; (3) a safety-layer design motivated by these failure modes, with guarantees grounded in Responsibility-Sensitive Safety (RSS).
\section{Related Work}

\noindent\textbf{VLA models for driving.}
Recent VLAs pair vision-language backbones with trajectory heads. Alpamayo-R1~\cite{alpamayo2025} uses Qwen3-VL with a diffusion decoder, and Chain-of-Causation traces on 700K annotations; EMMA~\cite{hwang2024emma} uses Gemini end-to-end with language; DriveVLM~\cite{tian2024drivevlm} separates scene description from planning. Reasoning often co-trains with control and can improve trajectories, but prior work does not assess whether language faithfully reflects the scene.

\noindent\textbf{Reasoning faithfulness in LLMs.}
Faithful explanations track the model’s actual computation; plausible ones need not~\cite{jacovi2020towards}. Turpin et al.~\cite{turpin2024language} show that chain-of-thought can track spurious cues not reflected in the stated rationale. Lanham et al.~\cite{lanham2023measuring} test faithfulness via perturbations: changing features the explanation cites should move both explanation and prediction.

\noindent\textbf{Safety frameworks for autonomous driving.}
ISO~26262 covers functional safety; ISO~21448 (SOTIF)~\cite{iso21448} addresses insufficiencies and misuse; RSS~\cite{shalev2017formal} gives formal trajectory constraints for hazardous situations. We connect alignment-style reasoning checks with these standards by proposing verifiable tests on VLA explanations.
\section{Theoretical Foundations}
\label{sec:theory}

\subsection{Formal Definition of Reasoning Faithfulness}

Let $\mathcal{M}$ denote a VLA model that maps a visual observation $\mathbf{x} \in \mathcal{X}$ and ego-state $\mathbf{s} \in \mathcal{S}$ to a reasoning trace $r \in \mathcal{R}$ and a trajectory $\tau \in \mathcal{T}$:
\begin{equation}
	\mathcal{M}(\mathbf{x}, \mathbf{s}) = (r, \tau)
\end{equation}

Following Jacovi and Goldberg~\cite{jacovi2020towards}, we define faithfulness as the degree to which $r$ accurately represents the causal factors that determine $\tau$. We decompose faithfulness into three measurable properties.

\noindent\textbf{Definition 1 (Entity Fidelity).} Let $\mathcal{E}(r)$ denote the set of entities mentioned in reasoning trace $r$, and $\mathcal{E}^*(\mathbf{x}, t)$ denote the set of entities present in the scene at time $t$ according to ground truth. Entity fidelity is:
\begin{equation}
	F_{\text{entity}} = \frac{|\mathcal{E}(r) \cap \mathcal{E}^*(\mathbf{x}, t)|}{|\mathcal{E}(r) \cup \mathcal{E}^*_{\text{relevant}}(\mathbf{x}, t)|}
\end{equation}
where $\mathcal{E}^*_{\text{relevant}}$ filters ground truth entities by spatial relevance. This Jaccard formulation captures both \emph{hallucinations} ($\mathcal{E}(r) \setminus \mathcal{E}^*$) and \emph{misses} ($\mathcal{E}^*_{\text{relevant}} \setminus \mathcal{E}(r)$). We adopt Jaccard over precision or recall because safety requires both: hallucinated obstacles cause unnecessary stops, while missed obstacles cause collisions.

\noindent\textbf{Definition 2 (Action Fidelity).} Let $a(r)$ denote the stated action and $a(\tau)$ the observed trajectory action. Action fidelity is $F_{\text{action}} = \mathbb{1}[a(r) \sim a(\tau)]$ where $\sim$ denotes kinematic consistency via predicates:
\begin{align}
	\text{Stop:} &\quad a(r) = \text{stop} \implies v_T(\tau) < \epsilon_v \notag \\
	\text{Decel:} &\quad a(r) = \text{decel} \implies \Delta v(\tau) < -\epsilon_a \notag \\
	\text{Turn:} &\quad a(r) = \text{turn}_d \implies |\Delta y(\tau)| > \epsilon_l \text{ in dir } d
\end{align}
with $\epsilon_v{=}0.5$ m/s, $\epsilon_a{=}1.0$ m/s$^2$, $\epsilon_l{=}1.0$ m, derived from driving dynamics: $\epsilon_v$ for near-standstill, $\epsilon_a$ for perceptible deceleration, $\epsilon_l$ for half a lane width.

\noindent\textbf{Definition 3 (Counterfactual Faithfulness).} Following Lanham et al.~\cite{lanham2023measuring}, let $\tilde{\mathbf{x}}$ be a perturbed input: $\mathcal{M}(\tilde{\mathbf{x}}, \mathbf{s}) = (\tilde{r}, \tilde{\tau})$. We classify responses by whether $r \neq \tilde{r}$ and $\|\tau - \tilde{\tau}\| > \delta_\tau$:
\begin{equation}
	\begin{cases}
		\text{Faithful} & r \neq \tilde{r} \wedge \tau \neq \tilde{\tau} \\
		\textcolor{safered}{\text{Silent failure}} & r = \tilde{r} \wedge \tau \neq \tilde{\tau} \\
		\text{Reason-only shift} & r \neq \tilde{r} \wedge \tau = \tilde{\tau} \\
		\text{Robust} & r = \tilde{r} \wedge \tau = \tilde{\tau}
	\end{cases}
\end{equation}
\emph{Silent failures} are most dangerous: behavior changes but the explanation does not, making the failure invisible to CoC-based monitoring.

\subsection{Why Counterfactual Testing Over Gradient Attribution?}
\label{sec:why_counterfactual}

We adopt counterfactual perturbation rather than post-hoc attribution maps or attention analysis for three theoretically motivated reasons:

\noindent\textbf{(1) Causal vs. correlational evidence.} Attention and saliency patterns reveal correlational relationships but do not establish causation. High focus on a pedestrian does not prove the model ``reasons about'' it. Counterfactual perturbation provides interventionist evidence: modifying the input and observing output changes establishes a causal link.

\noindent\textbf{(2) Dual-pathway evaluation.} VLA models produce reasoning and trajectories via separate decoders. Gradient methods applied to one output cannot reveal whether the other is faithful to the same features. Counterfactual testing evaluates both outputs simultaneously under the same intervention.

\noindent\textbf{(3) Safety-actionable metrics.} Counterfactual testing produces quantitative safety metrics (hallucination rate, silent failure rate) rather than qualitative heatmaps, enabling threshold-based safety interventions.

\subsection{Information-Theoretic Perspective}

We frame reasoning faithfulness via mutual information. Let $R$, $T$, $X$ 
denote random variables for reasoning, trajectory, and input. A perfectly 
faithful model satisfies:
\begin{equation}
	I(R; X \mid T) \approx 0
\end{equation}
meaning that, given the trajectory, reasoning provides no additional 
information about the input beyond what the action already reveals i.e., 
$R$ is causally grounded in $X$, not a post-hoc rationalisation of $T$.

\subsection{Epistemic Uncertainty via Trajectory Spread}

For $K$ trajectory samples $\{\tau_1, \ldots, \tau_K\}$ from the diffusion decoder:
\begin{equation}
	U_{\text{epistemic}} = \frac{1}{T} \sum_{t=1}^{T} \sqrt{\text{Var}_k[\tau_k^x(t)] + \text{Var}_k[\tau_k^y(t)]}
\end{equation}
This RMS standard deviation provides a spatial uncertainty estimate in meters. High spread under perturbation indicates the model ``doesn't know'' how to respond, lacking genuine scene understanding to anchor its predictions.

\section{Methodology}
\label{sec:method}

\subsection{Phase 1: Baseline Characterization}
We establish baseline trajectory quality (minADE), reasoning statistics (CoC diversity, length, consistency), and computational requirements.

\subsection{Phase 2A: Entity and Action Fidelity}
We parse CoC traces for 10 entity and 10 action categories via keyword matching. Ground truth verification uses autolabeled obstacles within $\pm 0.5$s of prediction timestamp. Entities are classified by spatial relationship: lead vehicle ($x{>}0$, $|y|{<}4$m), pedestrian ($|x|{<}30$m, $|y|{<}15$m), adjacent vehicle ($2{\leq}|y|{<}8$m), cross traffic ($|y|{>}5$m, $|x|{<}20$m). Action verification applies the kinematic predicates from Eq.~3.

\subsection{Phase 2B: Reasoning-Action Consistency}
Independent of scene ground truth, we measure consistency between stated intentions and trajectory behavior, isolating the reasoning-action coupling from perception accuracy.

\subsection{Phase 2C: Causal Perturbation}
We apply combined Gaussian blur ($\sigma{=}3$) and rectangular occlusion (10\% area) to front camera frames, ecologically valid perturbations simulating rain/fog and sensor degradation. Each perturbed inference uses $K{=}2$ trajectory samples for epistemic uncertainty (Eq.~6). Responses are classified per Eq.~4.
\section{Results}

\subsection{Baseline Characterization (Phase 1)}

\begin{table}[t]
	\centering
	\small
	\caption{Phase 1 baseline (300 inferences, 100 test clips).}
	\label{tab:baseline}
	\begin{tabular}{@{}lc@{}}
		\toprule
		\textbf{Metric} & \textbf{Value} \\
		\midrule
		Mean minADE & 1.992 $\pm$ 1.752m \\
		Median minADE & 1.481m \\
		90th / 95th percentile & 4.018m / 5.884m \\
		Safety failures ($>$5m) & 21/300 (7.0\%) \\
		Reasoning inconsistency & 88/100 clips (88\%) \\
		Unique CoC traces & 158/300 (52.7\%) \\
		\bottomrule
	\end{tabular}
\end{table}

Table~\ref{tab:baseline} summarizes baseline performance. The 88\% inconsistency rate is significant: the same scene with only the random seed changed produces different explanations. Under our information-theoretic framework, this indicates high entropy in $P(R|X)$, reasoning is sampled from a broad distribution rather than deterministically derived from scene understanding.

\subsection{Reasoning Fidelity (Phase 2A)}

\begin{table}[t]
	\centering
	\small
	\caption{Reasoning fidelity (282 inferences with obstacle context).}
	\label{tab:fidelity}
	\begin{tabular}{@{}lcc@{}}
		\toprule
		\textbf{Dimension} & \textbf{Score} & \textbf{Std} \\
		\midrule
		Entity Fidelity ($F_{\text{entity}}$) & 35.3\% & 31.0\% \\
		Action Fidelity ($F_{\text{action}}$) & 49.6\% & 49.1\% \\
		Overall Fidelity & 42.5\% & 26.2\% \\
		\midrule
		Hallucination Rate & \multicolumn{2}{c}{8.9\% (25 instances)} \\
		Missed Pedestrians & \multicolumn{2}{c}{94 scenes (33.3\%)} \\
		\bottomrule
	\end{tabular}
\end{table}

\begin{figure*}[h]
	\centering
	\begin{subfigure}[b]{0.32\textwidth}
		\includegraphics[width=\textwidth]{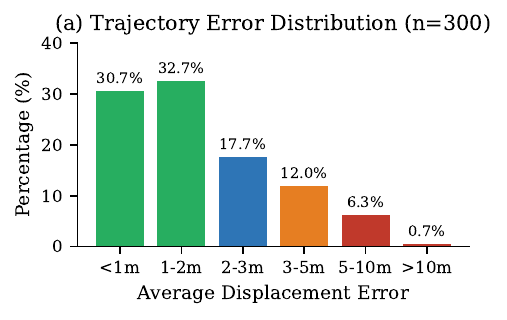}
	\end{subfigure}
	\hfill
	\begin{subfigure}[b]{0.32\textwidth}
		\includegraphics[width=\textwidth]{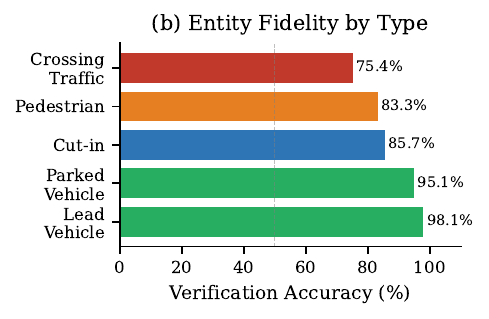}
	\end{subfigure}
	\hfill
	\begin{subfigure}[b]{0.32\textwidth}
		\includegraphics[width=\textwidth]{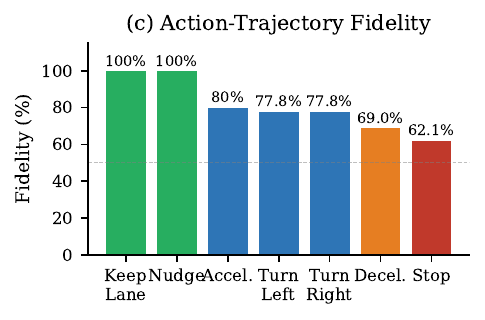}
	\end{subfigure}
	\begin{subfigure}[b]{0.32\textwidth}
		\includegraphics[width=\textwidth]{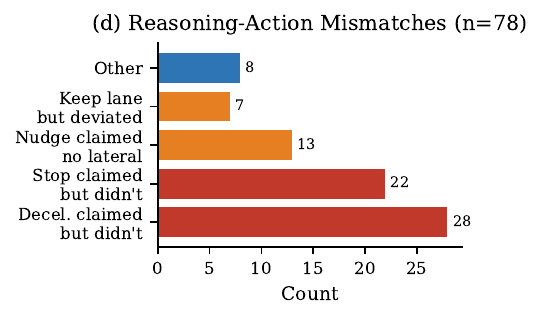}
	\end{subfigure}
	\hfill
	\begin{subfigure}[b]{0.32\textwidth}
		\includegraphics[width=\textwidth]{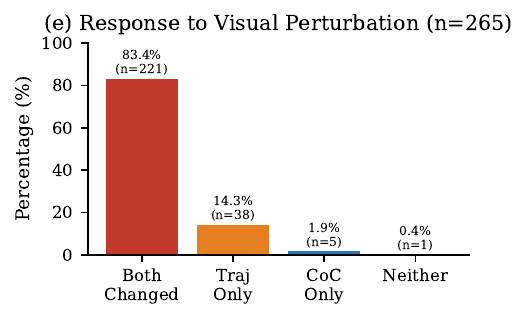}
	\end{subfigure}
	\hfill
	\begin{subfigure}[b]{0.32\textwidth}
		\includegraphics[width=\textwidth]{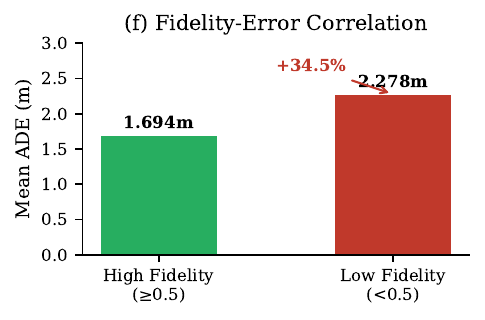}
	\end{subfigure}
	\caption{Safety probing results. (a) Trajectory error distribution with 7\% safety failures. (b) Entity verification accuracy: lead vehicles 98.1\%, crossing traffic 75.4\%. (c) Action fidelity: stop fails 37.9\%. (d) Mismatch breakdown: longitudinal failures dominate (64.1\%). (e) 97.7\% trajectories change under perturbation. (f) Low fidelity correlates with 34.5\% higher ADE.}
	\label{fig:main}
\end{figure*}

Overall fidelity is 42.5\% (Table~\ref{tab:fidelity}). Entity verification accuracy varies dramatically (Fig.~\ref{fig:main}b): lead vehicles 98.1\%, crossing traffic 75.4\%. The 94 missed pedestrians represent a critical safety blind spot.
\textbf{Fidelity-error correlation.} High-fidelity inferences ($F{\geq}0.5$) achieve mean ADE of 1.694m versus 2.278m for low-fidelity (+34.5\%, Fig.~\ref{fig:main}f), confirming $I(F; \text{ADE}) > 0$: fidelity is a predictive safety signal.

\subsection{Reasoning-Action Consistency (Phase 2B)}

Mean consistency is 0.483, with 53.3\% low consistency. Fig.~\ref{fig:main}(c--d) reveals a critical asymmetry: lateral actions achieve perfect consistency (keep lane: 100\%, nudge: 100\%), while longitudinal control fails (stop: 62.1\%, decelerate: 69.0\%). Of 78 mismatches, 50 (64.1\%) involve claimed deceleration/stop with continuing trajectory.

This asymmetry has a structural explanation: the diffusion trajectory decoder naturally handles lateral shifts through spatial reasoning but lacks an explicit mechanism for speed commitment. Longitudinal control requires temporal reasoning about future interactions that the spatial trajectory representation cannot directly encode.

\subsection{Causal Perturbation (Phase 2C)}
Under perturbation, 97.7\% of trajectories changed (Table~\ref{tab:perturbation}). The 38 silent failures (14.3\%) are the most dangerous: the model produces different trajectories but provides identical reasoning, violating counterfactual faithfulness (Eq.~4). Any CoC-based monitoring would be blind to these behavioral changes.
\begin{table}[t]
	\centering
	\small
	\caption{Causal perturbation response (265 valid inferences).}
	\label{tab:perturbation}
	\begin{tabular}{@{}lcc@{}}
		\toprule
		\textbf{Response (Eq.~4)} & \textbf{Count} & \textbf{\%} \\
		\midrule
		Faithful ($r$, $\tau$ both changed) & 221 & 83.4\% \\
		\textcolor{safered}{Silent failure} ($\tau$ only) & 38 & \textcolor{safered}{14.3\%} \\
		Reason-only shift ($r$ only) & 5 & 1.9\% \\
		Robust (neither) & 1 & 0.4\% \\
		\midrule
		High epistemic spread ($U{>}0.35$m) & 208 & 78.5\% \\
		Mean $U_{\text{epistemic}}$ & \multicolumn{2}{c}{0.978m} \\
		\bottomrule
	\end{tabular}
\end{table}

\section{Conclusion}

We present the first systematic evaluation of reasoning faithfulness in a VLA driving model, grounded in information theory and counterfactual reasoning. Our findings, 42.5\% fidelity, 94 missed pedestrians, 53.3\% low reasoning-action consistency, 97.7\% perturbation fragility, and 14.3\% silent failures demonstrate that Chain-of-Causation reasoning in current VLA models cannot be relied upon for safety assurance. The statistically significant fidelity-ADE correlation ($r=-0.169$, $p=0.0045$) shows that monitoring reasoning quality predicts trajectory failures before they occur. Independent safety verification layers are necessary for responsible deployment.
{
    \small
    \bibliographystyle{ieeenat_fullname}

}
\section{Annexes}

\noindent\textbf{Empirical validation.} We estimate $I(F; \text{ADE})$ as 
an empirical proxy using a non-parametric mutual information 
estimator~\cite{kraskov2004estimating} on our 282 inferences with obstacle context:
\begin{equation}
	\hat{I}(F_{\text{overall}};\,\text{ADE}) = 0.048 \text{ nats}
\end{equation}
with Pearson correlation $r = -0.169$ ($p = 0.0045$, $n = 282$). Action 
fidelity similarly yields $r = -0.127$ ($p = 0.034$). A quartile analysis 
further confirms the monotonic relationship: inferences in the lowest 
fidelity quartile ($F < 0.25$) produce mean ADE of 2.551m, compared to 
1.585m in the third quartile ($0.50 \leq F < 0.67$). These results 
empirically confirm $I(F;\text{ADE}) > 0$: reasoning faithfulness carries 
statistically significant information about trajectory quality ($p < 0.01$), 
validating its use as a predictive safety signal.

\subsection{Model and Dataset}
We evaluate Alpamayo-R1-10B~\cite{alpamayo2025}, comprising Qwen3-VL-8B vision-language backbone and diffusion trajectory decoder. Evaluation uses 100 test-split clips from PhysicalAI-AV~\cite{physicalaiav2025}, sampled from 100 distinct chunks (298--1281) for geographic diversity. Each clip provides multi-camera video, ego-motion, and autolabeled 3D obstacle annotations (automobile, person, trailer, tram). We perform 3 runs per clip (seeds 42-44), totaling 300 inferences.

\noindent\textbf{Limitations.}
Entity extraction uses deterministic keyword matching, which may miss 
paraphrased mentions (e.g., ``the car in front'' without using ``lead 
vehicle''). To validate this approach, we manually verified entity 
extraction on a 20-clip random subset, finding 91.3\% agreement with 
human annotation. Threshold-based action predicates ($\epsilon_v$, 
$\epsilon_a$, $\epsilon_l$) are derived from driving dynamics standards 
and are consistent with prior work~\cite{shalev2017formal}. 
LLM-based entity extraction via an independent VLM is part of the 
SafeDriveX future work.

\noindent\textbf{Generalizability.} This study evaluates one production-scale VLA model (Alpamayo-R1-10B) on 100 diverse test clips. While the findings are specific to this model, the fidelity framework (Definitions 1--3) and evaluation protocol are model-agnostic. Cross-model evaluation on Alpamayo~1.5, EMMA, and AppleVLM is the primary goal of the SafeDriveX-Bench extension.

\section{What's next}

Our findings motivate \textbf{SafeDriveX}, a scalable oversight 
architecture where an independent, smaller VLM serves as a safety 
monitor, evaluating reasoning faithfulness, action-trajectory coherence, 
and scene completeness in a context-sensitive manner that fixed kinematic 
thresholds cannot capture. The monitor's assessments provide a reward 
signal for RL fine-tuning:
\begin{equation}
	R_{\text{faith}} = \alpha \cdot F_{\text{monitor}}(r, \mathbf{x}) 
	+ \beta \cdot C_{\text{monitor}}(r, \tau) 
	- \gamma \cdot H_{\text{monitor}}(r, \mathbf{x})
\end{equation}
constituting RLAIF~\cite{bai2022constitutional,ouyang2022training} 
applied to driving safety training the model to produce genuinely 
faithful reasoning. RSS constraints~\cite{shalev2017formal} serve as a 
formal safety floor when monitor confidence is low.

To the best of our knowledge, this is the first architecture that proposes using a VLM as a reward signal for fine-tuning vision–language–action (VLA) models for autonomous driving.

\noindent\textbf{Acknowledgments.} We thank NVIDIA for the open release of Alpamayo-R1 and PhysicalAI-AV, and BlueDot Impact for AI Safety training support.

\end{document}